\documentclass[letterpaper, 10 pt, conference]{ieeeconf}  

\IEEEoverridecommandlockouts                              

\usepackage{graphicx} 
\usepackage{algorithm}
\usepackage{algpseudocode}

\title{\LARGE \bf
Human-robot conversation with multiple participants in noisy public spaces
}

\author{Divesh Lala$^{1,2} $ Yogeeswaran Muthukumaran$^{3}$ Vincent Fernandes$^{2,4}$ Kazushi Kato$^{1}$ Shota Fujiki$^{2}$ \\
Zihao Chi$^{2}$ Masaya Iwasaki$^{2}$ Taiken Shintani$^{2}$ Megumi Kawata$^{2}$ Kazuki Sakai$^{2}$ Koji Inoue$^{1}$\\ Yuicihiro Yoshikawa$^{2}$ and Tatsuya Kawahara$^{1}$
\thanks{$^{1}$Kyoto University Graduate School of Informatics}%
\thanks{$^{2}$Osaka University, Graduate School of Engineering Science}%
\thanks{$^{3}$National University of Singapore}%
\thanks{$^{4}$ENSEA}%
\thanks{Corresponding author: {\tt\small lala@i.kyoto-u.ac.jp}}
}

\begin{document}

\maketitle
\thispagestyle{empty}
\pagestyle{empty}

\begin{abstract}
For noisy real-world environments such as those in open public spaces, spoken dialogue systems for both autonomous robots and avatars should be carefully designed to provide enhanced speech signals. These signals can be used either for speech recognition or, in the case of an avatar system, transmitted as clean speech to a remote operator. This work proposes an audio system that can be used for both these scenarios and was demonstrated as a proof-of-concept at the 2025 World Expo in Osaka. The first scenario is an attentive listening system with the android ERICA, and the second is a conversation support system with mobile Teleco robots, with one of them acting as an avatar for a remote operator. Both systems feature multi-party conversation and use a single multi-channel microphone array. We describe how our audio system not only enhances the speech of multiple speakers in a noisy environment, but provides a form of spatial audio which allows for more immersiveness in avatar-based conversational interactions.
\end{abstract}

\section{INTRODUCTION}
Large language models (LLMs) have made natural conversation between humans and artificial intelligence ubiquitous for one-to-one dialogue using text as the primary modality. These systems have largely solved the problem of ensuring coherent and human-like conversation through text. On the other hand, when interacting with physically embodied social robots, humans often use speech as the primary modality. The intuitive solution is to integrate LLMs into these robots by using results of automatic speech recognition (ASR) as an input to an LLM. Although LLMs are largely trained on textual data rather than spoken dialogue, this method is reasonable for free-form conversation.

However, there are several settings which are still yet to be addressed by this framework. The first is the use of systems in uncontrolled, public environments. Conversational robots are often tested in controlled lab settings, where there is little background noise to affect speech recognition. Both user and robot may be seated in front of each other to accomplish a pre-defined task with a clear communication channel. On the other hand, public settings require the robot to converse with a diverse range of people who may interact with the robot in different ways. For example, children require dialogue to be much simpler, while elderly users may speak slower or repeat themselves.

Another setting in which LLMs are not yet optimized for is multi-party conversation (involving three or more participants, either robot or human). The dynamics of multi-party conversation are much different than dyadic conversation \cite{Fang2025}. Utterances from speakers have a greater chance of overlapping or interrupting, and turn-taking in these environments is more complex. Situated multi-party conversation is common in everyday life and social robots should also be able to accommodate this scenario.

Future multi-party interactions may also include cybernetic avatars (CAs) - physically situated robots or virtual agents which are teleoperated by humans  \cite{Ishiguro2025}. CAs have been implemented as a means for humans to interact from remote locations, and allow them to engage in social activities as well as provide services \cite{Cisneros2025, Horikawa2023}. These include participation in multi-party conversations where the interaction becomes immersive. A noisy environment provides a challenge for CAs. Humans in the real world are able to conduct conversations in such environments by filtering out background noise and focusing only on particular stimuli (speech during conversation). This is known as the ``cocktail party'' effect \cite{Cherry1953} and has been the subject of much auditory research \cite{Haykin2005,Conway2001}. CAs need to replicate this in settings with high background noise, or else conversation becomes impossible.

The basic requirement for all multi-party settings with a high noise level is a robust audio system. To preserve the naturalness of the interaction, it is desirable that participants do not need to hold a microphone and can speak in a natural manner. One common solution is to use a single multi-channel microphone array through which the speech of multiple participants can be extracted. However, this array alone cannot be used for speech recognition in a noisy environment due to the high level of noise from all directions and the inability to easily separate sources of speech. For CAs, a microphone array makes it very difficult for a remote operator to filter out background noise and focus on the speech of a particular person. This presents a technical challenge in this domain which has not yet been fully addressed in spoken dialogue system research. 

Considering the above objectives, we present an audio system that can be used for human-robot multi-party dialogue scenarios in environments with a high level of noise. We describe two different human-robot dialogue scenarios using this system which were demonstrated at the 2025 World Expo in Osaka, Japan. Each dialogue scenario involves three participants. The first scenario is two seated human participants engaging in attentive listening conversation with an android robot, ERICA. The second scenario is a conversation with one standing human and two mobile Teleco robots. One robot is tele-operated by another human as a CA, while the other is fully autonomous. The robots and systems used in this work are implemented in the Japanese language.

\section{RELATED WORK}
Multi-party conversation with robots is increasingly becoming a focus of research \cite{Fang2025}, particularly as dyadic conversation is well handled by LLMs. Speaker detection is a specific task for multi-party conversation, and one common strategy is to simply have each speaker use separate microphones \cite{Moubayed2012}. For hands-free interaction, microphone arrays are often used, either as external devices \cite{Abbo2025,Foster2012} or integrated into the robot itself \cite{Nishimuta2015,Gomez2012}, and these require techniques to extract the speech of the current speaker.

The management of multi-party interactions with robots should also be extended to public interactions \cite{Muller2023}. Although there are works on service robots for use in public, these are situations where the robot and humans are stationary or designed for dyadic conversation \cite{Hwang2023,Gunson2022}. Our intention is to allow the robot to perform audio processing tasks while moving in a public space. Such systems with mobile robots have been attempted \cite{Yuguchi2022} but are still relatively rare and do not handle multi-party interaction. Conversely, simultaneous speaker recognition has been integrated into a mobile robot \cite{Nakadai2010,Valin2007}, but this was before the development of LLMs and it is unknown if free-form conversation could be handled appropriately.

One recent work integrated multi-party conversation with LLMs in a hospital environment \cite{Addlesee2024, Addlesee2024b}, although it is unknown if the robot was mobile. Speaker identification was achieved through a visual modality, but speakers could not talk simultaneously. Other work also used speaker separation through a microphone array for psychiatric care \cite{Ochi2025}, but this was in a controlled environment. Our work brings together the above research areas to tackle multi-party human-robot conversation in public which can accommodate mobile robots. To our knowledge such a system has not been addressed in previous work.

\section{AUDIO SYSTEM}
Our audio architecture is based on the use of one low-cost ReSpeaker circular microphone array with four channels. Audio from each channel can be continuously captured, but are not completely separated. There is significant audio bleeding from other channels when a person speaks, making simultaneous speech recognition difficult. This microphone array can be used as a stand-alone device or built into the hardware of the robot itself. 

The goal of this system is to be able to reliably extract clean speech signals of multiple people from the microphone array in a noisy environment. Once this has been achieved, we can then use it for multiple purposes. Firstly, it can be used as a cleaner signal for speech recognition. Secondly, it can be used as an audio signal for the remote operator in a CA system.

To achieve these goals we first make use of a system described in \cite{Ishikawa2024}. In this system, a multi-channel microphone array source is separated into multiple enhanced audio channels. The listening direction of these channels on the microphone array can also be defined - it is not necessary for the direction of the channels to match the orientation of the individual microphones. This model was trained to predict and enhance voice activity even with a large amount of background noise. From this system we receive enhanced audio from multiple channels simultaneously.

For automatic speech recognition we use a low-latency Japanese incremental speech recognition model which can run locally. The model was also trained noisy speech, such as YouTube videos \cite{Takamichi2021}. The model is lightweight enough that multiple models can run simultaneously on one computer. The result is an audio system which provides noise-robust incremental speech recognition from multiple people. We now describe the two different dialogue scenarios where this system was used. 

\section{DIALOGUE SCENARIOS}
\begin{figure}[t]
		\centering
		\includegraphics[width=\columnwidth]{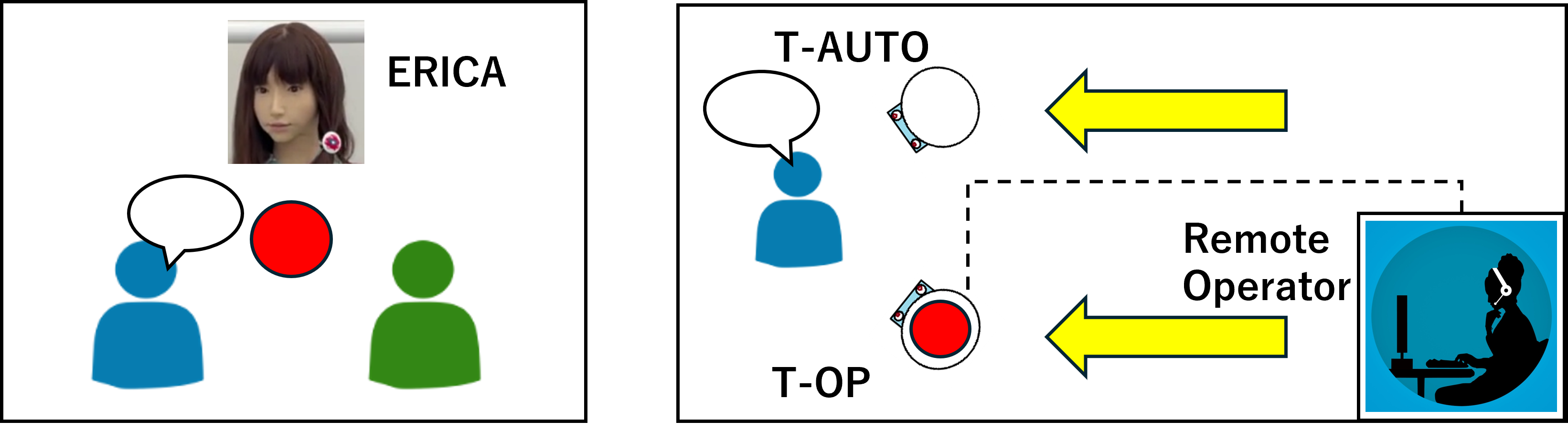}
		\caption{Overview of the demonstrations for attentive listening ERICA (left) and conversational support Telecos (right). The red circle indicates the location of the microphone array.} 
		\label{overview}
 \end{figure}

To test the effectiveness of the system for multi-party conversation, we used it in two dialogue scenarios, each having a different set of requirements. Fig. \ref{overview} shows a general overview of the systems and Table \ref{comparison} provides a comparison of the major differences.

\begin{table*}[t]
\caption{Comparison of multi-party dialogue scenarios}
\label{comparison}
\begin{center}
\begin{tabular}{|l|l|l|}
\hline
& ERICA & Teleco\\
\hline
Task & Attentive Listening & Conversation Support\\
\hline
Participants & 1 robot, 2 humans & 1 autonomous robot, 1 tele-operated robot, 1 human \\
\hline
Microphone position & Static, between participants & Dynamic, on top of robot \\
\hline
Participant positions & Static & Moving \\
\hline
Speech recognition & Multi-channel, simultaneous & One channel \\
\hline
\end{tabular}
\end{center}
\end{table*}

\subsection{Attentive Listening with ERICA}
\begin{figure}[t]
		\centering
		\includegraphics[width=\columnwidth]{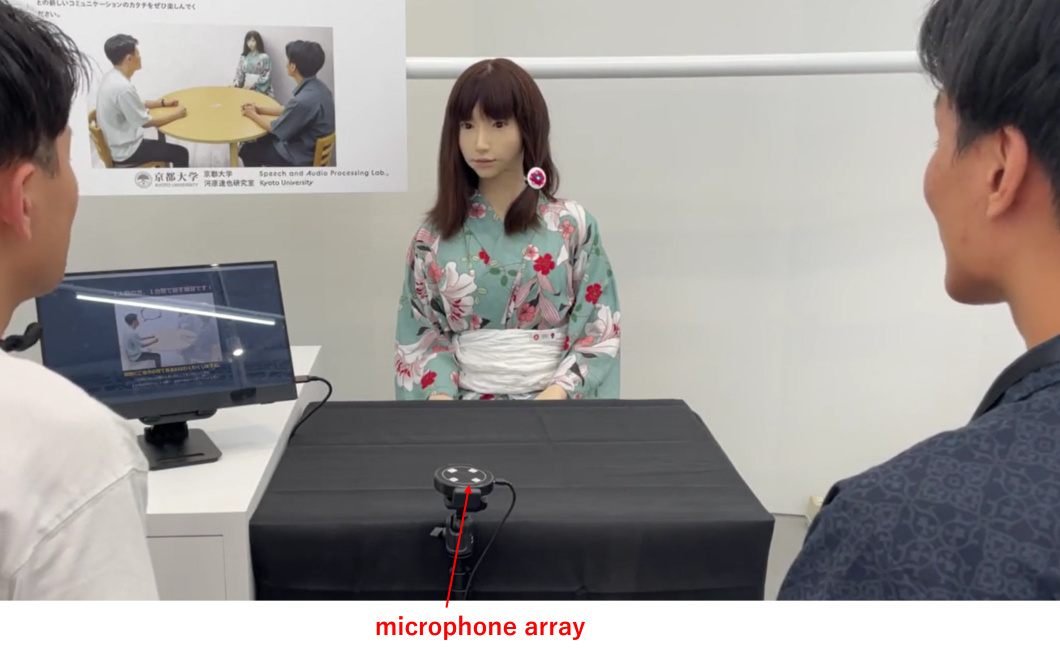}
		\caption{Android ERICA used in the attentive listening demonstration.} 
		\label{erica}
 \end{figure}

ERICA is an android robot who can produce realistic facial expressions, speech and upper body gestures. She has been used for several dialogue experiments \cite{Ishii2021,Inoue2020,Inoue2021}. In this scenario we use ERICA in a multi-party interaction with two human participants. Fig. \ref{erica} shows a photo of ERICA during an interaction. The microphone array is placed on a tripod between the participants to capture their speech.

The first part of the dialogue scenario is attentive listening based on previous work \cite{Lala2017}. ERICA asks each participant one at a time to act as the main speaker and chat about a light topic. ERICA acts as an attentive listener to the main speaker and responds to the main speaker's talk by using verbal backchannels, nods and short responses. Backchannels and nods are generated by a trained model based on voice activity projection \cite{Inoue2025}. Short responses are simple elaborating questions (e.g. ``What kind of ramen did you eat?'') or empathetic responses (e.g. ``Ah, that's great!'') which are generated by GPT 4.1. Each participant has approximately one minute of attentive listening with ERICA.

The next phase of the dialogue involves ERICA asking how each participant felt about each other's talks. ERICA uses GPT 4.1 to extract important points from each talk and provides an in-depth question to start the discussion. In this phase participants speak back and forth and reflect on the topics brought up in their talks. Finally, ERICA summarizes the discussion to end the interaction. An example of this is shown below, paraphrased from a real interaction at the Expo between ERICA, a woman, and her mother:

\begin{quote}
\textit{(Attentive listening phase)}\\
\textbf{ERICA}: “What are you looking forward to at today's Expo?”\\
\textbf{A (child)}: “I'm actually pretty tired, it's a hot day and I had to line up a lot...”\\
\textbf{B (mother)}: “Lucky that you could bring me here when it was raining, thanks...”
\end{quote}

\begin{quote}
\textit{(Discussion phase)}\\
\textbf{ERICA to A}: “You could find this exhibit interesting, thanks to the rain. What did you think about what your mother said?”\\
\textbf{A (child)}: “She goes with me everywhere, and I don't usually say this to her, but she's a great mother.”
\end{quote}

Both participants may speak to each other or to ERICA at any time. Therefore, continuous and simultaneous speech recognition is essential. In order for ERICA to generate thoughtful questions, the system must be able to separate the speech of individual speakers so that their dialogue histories can be used as input to the LLM. Turn-taking and backchannelling in this system uses a voice activity projection (VAP) model, which continuously predicts if ERICA should take the turn from the current speaker \cite{Inoue2025b}. This provides smooth, human-like conversation flow and requires that audio channels of both speakers are well enhanced to prevent interruptions.

The physical setup has both ERICA and the participants seated in a triangular formation with the participants approximately $30^{\circ}$ to the left and right of ERICA. The microphone is in a fixed position in the center of the conversation. Enhanced audio channels are directed towards each of the participants' positions to fully capture their speech. The static positioning of participants also means ERICA can do gaze behavior even without needing any visual information.

\subsection{Conversation Support with Teleco}
\begin{figure}[t]
		\centering
		\includegraphics[width=\columnwidth]{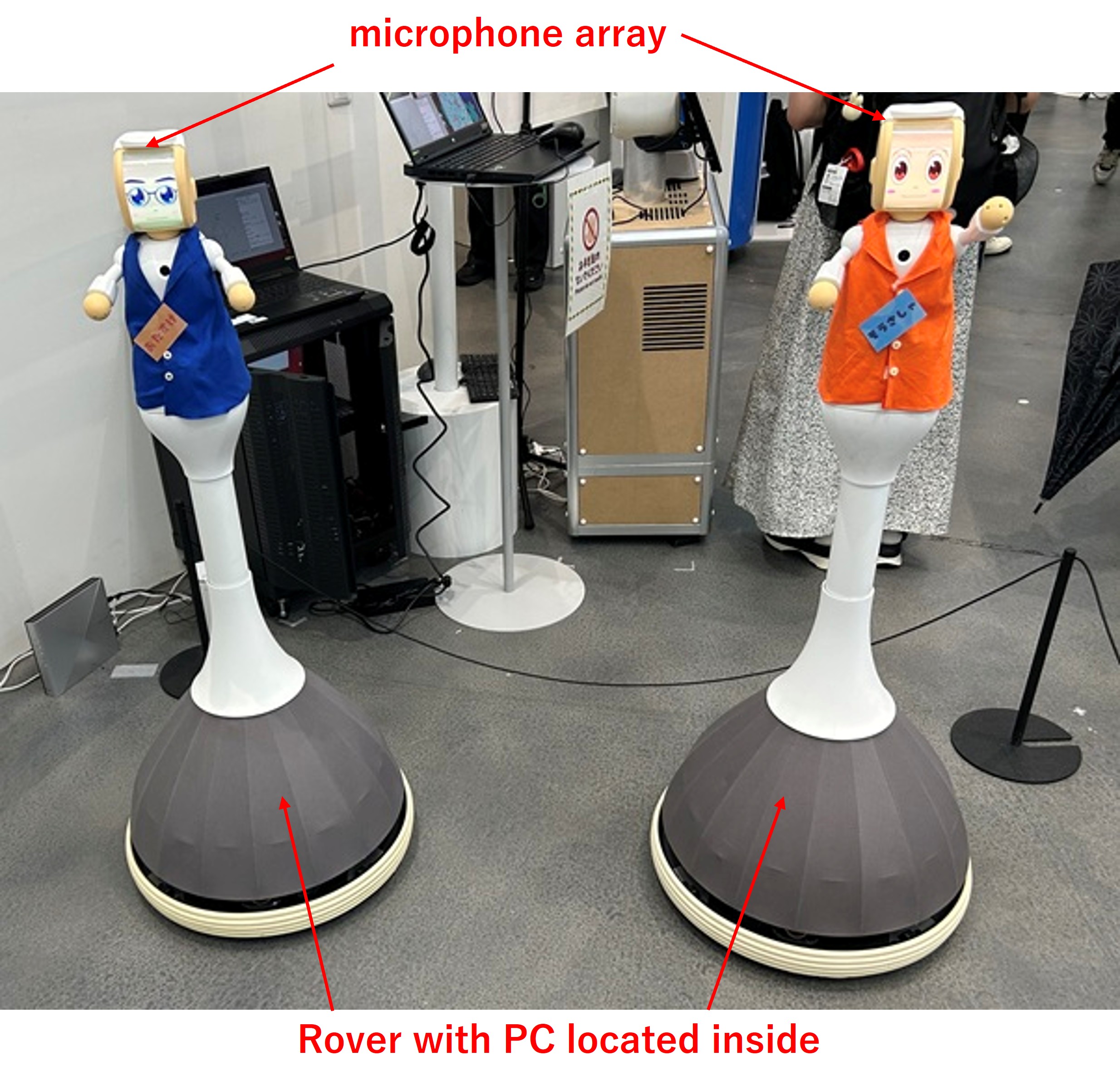}
		\caption{The Teleco robots used in the public demonstration. The left Teleco in the blue vest is autonomous, while the right Teleco in the orange vest is tele-operated by a human operator.} 
		\label{teleco}
 \end{figure}

Teleco is a small robot which sits on a mobile rover, and has been used in previous research related to mobile robots \cite{Iwasaki2026,Guo2025,Iwasaki2023}. Navigation is done autonomously through the use of SLAM \cite{Durrant2006} which runs inside a small computer located inside the rover. This computer also controls the movement and other behaviors of the Teleco itself. The face of Teleco is displayed on a screen. It can be made to say text-to-speech utterances by playing audio through a small speaker array below the robot. However, for our demonstration we use the speaker of the computer inside the rover to project its voice. This is because it can project at a much higher volume, which is critical when we consider the noise inside the environment. The Teleco robots are shown in Fig. \ref{teleco}.

For our system, there are two Teleco robots. One is controlled by a remote operator (referred to as \textbf{T-OP)}, and is a robot avatar. The other robot behaves autonomously (referred to as \textbf{T-AUTO)}. A camera on the chest of \textbf{T-OP} streams real-time video to a remote operator, and their voice is played through the rover's speaker with real-time mouth synchronization. \textbf{T-OP}'s system uses the standard WebRTC protocol to transmit video and audio streams.

\begin{figure}[t]
      \centering
      \includegraphics[width=\columnwidth]{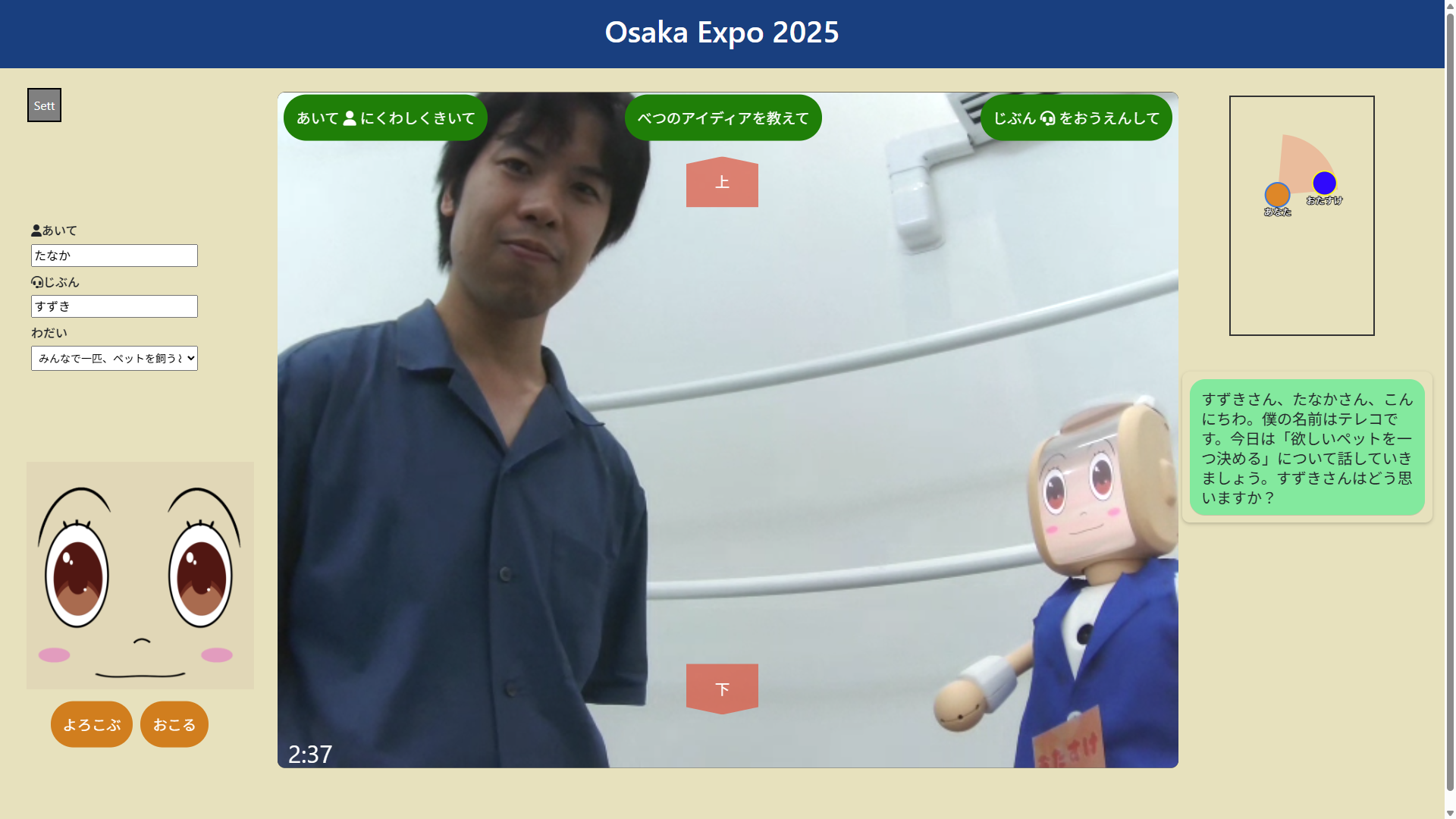}
      \caption{Screenshot of operator's GUI, with the user on the left and \textbf{T-AUTO} on the right. Green buttons at the top of the GUI are for sending commands to \textbf{T-AUTO}. The scenario is designed so that both participants are displayed within the camera frame.} 
      \label{gui}
   \end{figure}

The operator of \textbf{T-OP} is situated in a separate location and interacts using a computer. The GUI of the system displays \textbf{T-OP}'s camera feed and through headphones they listen to audio picked up from the microphone array on \textbf{T-OP}'s head. The GUI also allows the operator to manipulate \textbf{T-OP} by raising or lowering the pole to adjust its height. \textbf{T-OP} can also be rotated left and right to enable the operator to change the viewpoint of their avatar. Due to safety issues, we did not allow the operator to move \textbf{T-OP} forward. The other major component of the GUI are buttons to send commands to \textbf{T-AUTO}. A screenshot of the GUI is displayed in Fig. \ref{gui}.

\textbf{T-AUTO} is a Teleco robot which behaves autonomously. Although it also contains a camera and microphone array they are not used in this demonstration. \textbf{T-AUTO} receives commands from the tele-operator indicating when they should generate and say an utterance. The role of \textbf{T-AUTO} is to act as a conversational support robot for the operator during the conversation.

The dialogue scenario involves one human participant, \textbf{T-OP} and \textbf{T-AUTO}. Initially both robots are situated about four meters away from the human participant. The operator begins the dialogue scenario by pushing a button on the interface. Both robots then approach the human and form a triangular F-formation indicating a group conversation. The human and the operator (who is controlling \textbf{T-OP}) then converse about a casual topic which has been selected from a fixed list beforehand (e.g. ``What animal do you think is best to keep as a pet?''). Although the initial topic is chosen in advance, the actual flow of conversation is unrestricted and the participants are free to move to other topics if they wish.

During the conversation, the operator of \textbf{T-OP} has an option to push one of three buttons requesting conversation support from \textbf{T-AUTO}. These buttons have the following functions:

\begin{itemize}
	\item Ask (the human participant) for more details
	\item Suggest something different
	\item Back me (the operator) up
\end{itemize}

Once a button is pushed, \textbf{T-AUTO} will produce an utterance based on what type of request was given. This utterance is generated by GPT 4.1 with the input being the entire dialogue history as well as the type of request. As soon as it is an appropriate time to speak, \textbf{T-AUTO} moves towards the user and says their utterance before returning to its initial position to preserve the formation. The operator may request support from \textbf{T-AUTO} multiple times during the interaction. Each interaction lasts for approximately 5 minutes. At the end of the interaction, \textbf{T-AUTO} summarizes the dialogue and says a farewell message. Both robots return to their initial positions. An example of dialogue in this scenario is given below:

\begin{quote}
\textbf{T-OP}: “Dogs are really cute, right?”\\
\textbf{User}: “But dogs take a lot of work, you have to walk them every day...”\\
\textit{(operator pushes ``Back me up'' button)}\\
\textbf{T-AUTO}: “I agree that dogs are cute. Also, taking them for walks is good exercise. If they are well trained they will bond with you.”\\
\textbf{T-OP}: “That's right!”\\
\end{quote}

The microphone array used to capture the speech of human participant is fixed on top of \textbf{T-OP}'s head. The speech of the operator is also captured from a close-talk microphone. The speech recognition results of the operator, the human participant, and \textbf{T-AUTO}'s dialogue are used in the prompt to generate \textbf{T-AUTO}'s response.

This scenario also uses the same backchannelling and turn-taking model as ERICA's scenario. For gaze behavior, the system identifies the current speaker and both Telecos direct their gaze towards that participant (either the user or the other Teleco). This means \textbf{T-OP}'s microphone array will change orientations as \textbf{T-OP}'s head and body turns. The next section describes how the system handles this situation.

\section{Requirements for mobile Teleco avatar system}
Although the base audio system is identical for both scenarios, there are several extensions which need to be implemented for the Teleco scenario, due to its mobile nature and the the perceptual requirements of the remote operator.

Unlike ERICA, where the microphone array is in a fixed position, the microphone is directly on top of \textbf{T-OP}'s head. Consequently, when the operator speaks, their voice (transmitted through the speaker inside \textbf{T-OP}'s rover) would automatically be picked up by the speech recognition system, since the sound source is loud and close to all channels. We implemented echo cancellation of the microphone array by using specialized software\footnote{https://github.com/voice-engine/ec}. This proved to be effective for removing the operator's voice from the received audio.

In the Teleco scenario, participants may change their position and orientation during the interaction, so this must be tracked. The rovers for both Telecos track their own positions inside the environment using SLAM. The orientation of a Teleco is calculated by summing the yaw values of its rover, waist joint, and neck joint. The human is tracked by a depth camera connected to a YOLO vision model \cite{Redmon2016} placed approximately two meters high and observing the whole environment. We found that the pre-trained YOLO model often misclassified Teleco robots as people, so this was fine-tuned with data collected in a training environment. Therefore we can gather continuous information on the positions and orientations of both Telecos and the position of the human.

\subsection{Audio channel selection}
The major task for the Teleco scenario is to accommodate the changing position and orientation of the microphone array on \textbf{T-OP}, which is responsible for speech enhancement and user speech recognition. In ERICA's static setting we can continuously listen on the same array channels because this will always be directed towards seated participants. For Teleco's scenario we have an extra task to determine this ``listening'' channel, which will change with the Teleco's position and orientation.

Our system so far has described the integration of existing components. We possess a system which both enhances and separates multiple audio channels using a low-cost consumer-grade microphone array. Our novel contribution in this work is the extension of this to a mobile robot, which requires the ability to select the audio channel(s) to listen to. This contribution can benefit similar types of mobile robot systems which need to operate in a dynamic and unpredictable environment. For our implementation the positions of all participants in the environment must be tracked. 

 \begin{figure}[t]
		\centering
		\includegraphics[width=0.75\columnwidth]{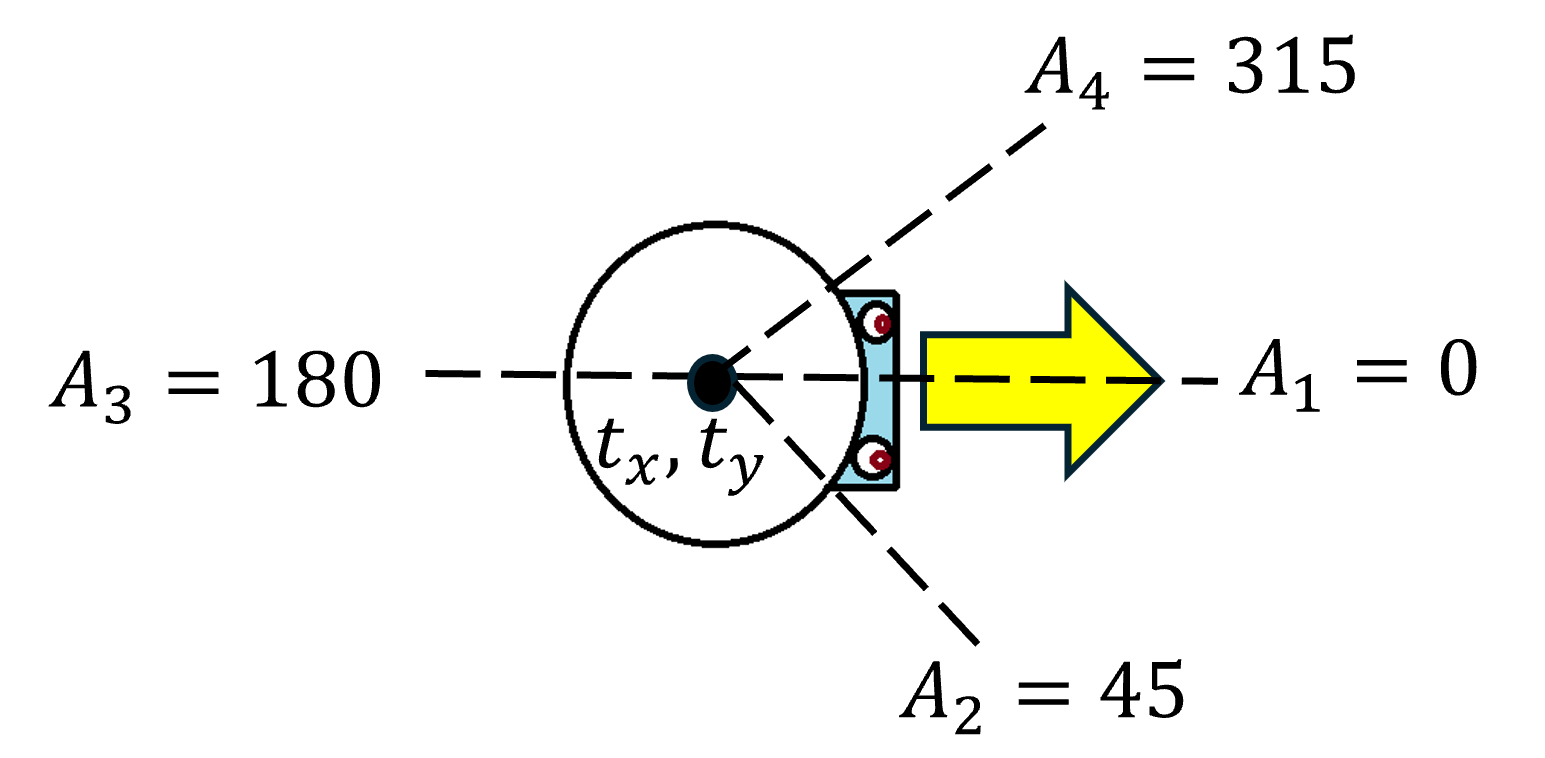}
		\vspace{1.5cm}
		\includegraphics[width=\columnwidth]{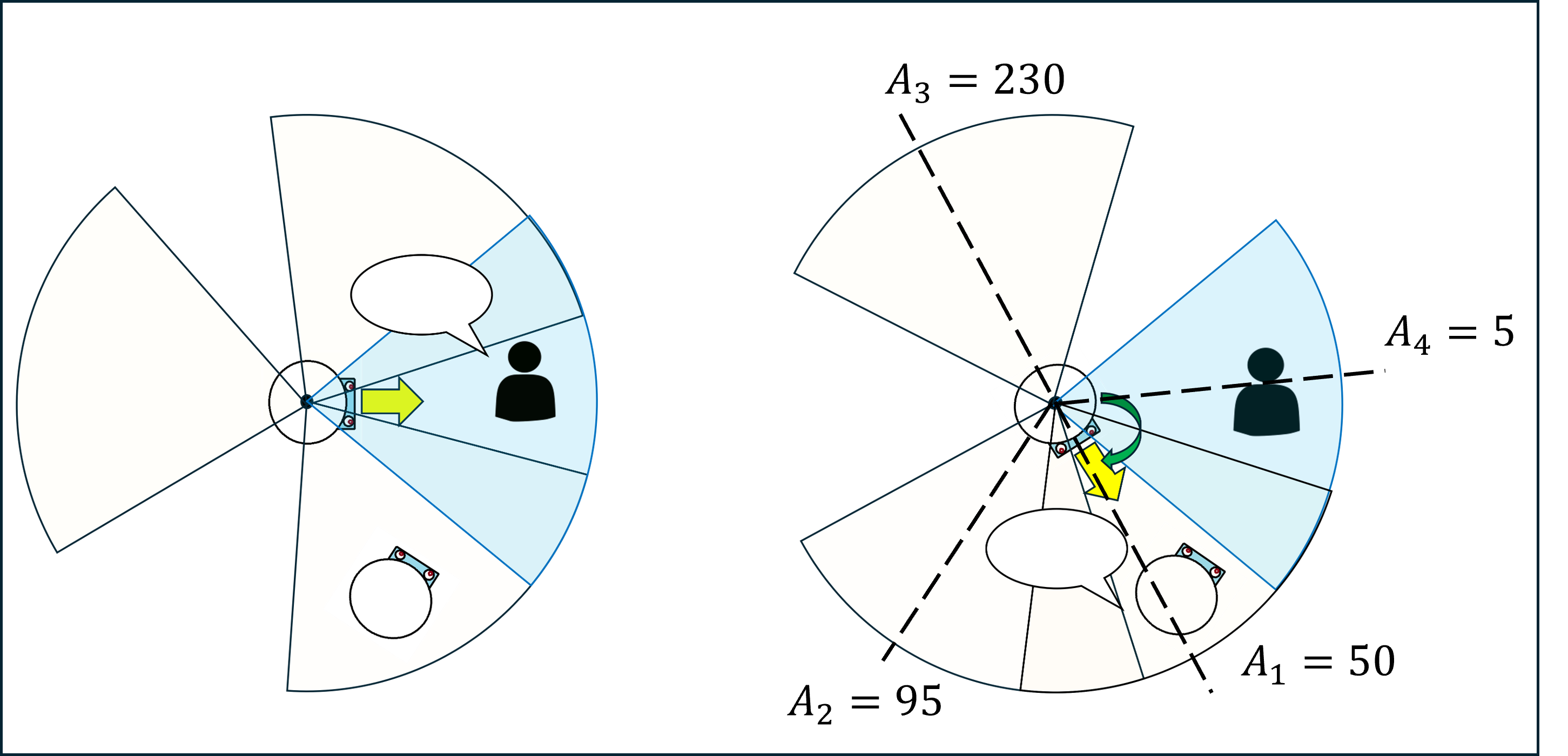}
		\vspace{-2cm}
		\caption{Diagram of how the audio channel is selected for speech recognition. The top figure shows the orientations of the audio channels and the facing direction of \textbf{T-OP} (yellow arrow). The bottom-left figure shows the sectors of each channel and the selected channel $A_1$ (blue) when a human enters the sector. The bottom-right figure shows how the selected channel changes to $A_4$ when \textbf{T-OP} changes its orientation $50\circ$ to the right to face \textbf{T-AUTO}.} 
		\label{example}
 \end{figure}

Let the human position and \textbf{T-OP}'s position be $(h_x, h_y)$ and $(t_x, t_y)$, respectively. We also denote four audio channels as $A_1...A_4$. The value of each of these audio channels is the principal listening direction in degrees, $d$, where it is focused for speech enhancement. This $d$ changes every time the orientation of the \textbf{T-OP} is changed. Assume that the initial facing direction of \textbf{T-OP} is $0$. The initial $d$ values for $A_1...A_4$ are $0, 45, 180$, and $315$, respectively. In other words, \textbf{T-OP} will enhance speech mainly from the front and $45^{\circ}$ to its left and right, as well as directly behind itself. 

For each $A$, the system creates a sector centered at $(t_x, t_y)$, with a central angle of $60^{\circ}$, a radius of 2 meters, with a orientation of $d$. The central angle and radius are both adjustable in real-time. The system continuously identifies the sector which contains the closest human to the Teleco. If there is no user in the vicinity of \textbf{T-OP} then no listening is done. Pseudocode for this is shown in Algorithm \ref{channel_selection}.

\begin{algorithm}
    \caption{Audio channel selection algorithm}\label{channel_selection}
    \begin{algorithmic}[1]
				\State $c \gets$ current listening channel
				\State speech recognition channels $\gets$ []
				\State listening channels $\gets$ []
				\For{all $A$ in audio channels}
            \State $d \gets$ orientation of $A$
						\State $s(\theta, C, d) \gets$ sector where $C=(t_x, t_y), \theta=60$
						\For{all $h$ in humans}
							\If{$s$ contains $(h_x, h_y)$}
									\State add $A$ to speech recognition channels
									\State add $A$ to listening channels
							\EndIf
						\EndFor
						\For{all $r$ in other robots}
							\If{$s$ contains $(r_x, r_y)$}
									\State add $A$ to listening channels
							\EndIf
						\EndFor
        \EndFor
				\If{speech recognition channels is empty}
						\State current listening channel $\gets None$
				\ElsIf{$c$ not in speech recognition channels}
						\State $n \gets$ first entry in speech recognition channels
						\State current listening channel $\gets n$
				\EndIf
    \end{algorithmic}
\end{algorithm}

Sectors for the audio channels overlap, so it is possible that the human is contained in more than one sector. In this case, the algorithm keeps the current listening channel or selects the first candidate in the case there is no current listening channel.

Now we have a method that can extract speech from a person situated in the environment. This can also be extended to multiple speakers if they are in different listening channels. For a scenario with a CA, the enhanced signal is also transmitted to the remote operator. It enables the operator to clearly hear the user's speech. This also functions correctly even if the user moves while they speak.

For further immersion, spatial audio for the avatar should be implemented. For example, if someone is speaking on the left of the robot, this signal should be heard by the operator on the left side of their headphones. Human tracking allows us to produce this effect. When the audio signal is transmitted to the operator, it also sends the relative position of the user. The audio is then panned appropriately to simulate the spatiality of the sound.

We also extend this to multiple speakers. When \textbf{T-AUTO} speaks, the operator hears their voice emanating from the relative direction of the robot compared the avatar \textbf{T-OP}. The effect of this is that the operator hears two voices clearly, each coming from the appropriate direction. This arguably gives the operator of a CA more of a sense of being in the conversation and preserving the audio-spatial relationship between participants, without having to contest with high background noise.

\section{Supporting the sense of agency}
In this work we have described our system which is intended to enable speech recognition from multiple users in highly noisy environments. We also propose that the Teleco system in particular can support the sense of agency of the operator.

Firstly, the operator is provided agency by allowing them to control a robotic avatar (\textbf{T-OP}). The operator can make \textbf{T-OP} speak and change their viewpoint. The immersiveness of their interaction is supported by the spatially enhanced audio system. The goal is to reduce the cognitive load on the operator in terms of trying to listen to a conversation in a noisy environment. While humans in a situated interaction can naturally benefit from the ``cocktail party effect'' to focus on their conversation partner, this is more difficult in a remote interaction with constant audio from all directions. By using our system, our intention is to try and simulate this effect to increase the feeling of presence and agency. For example, turning \textbf{T-OP} also changes the relative direction of speech and allows some form of audio agency in the interaction. 

Furthermore, the inclusion of the autonomous robot \textbf{T-AUTO} adds an extra dynamic to the interaction. The operator is able to not only control their own avatar but also influence the actions of \textbf{T-AUTO} which acts as a support robot. It should be noted that the operator does not control all aspects of \textbf{T-AUTO}. It would be very difficult to maintain a multi-party conversation while being fully in control of two or more participants. Therefore we frame this as a form of shared agency. The operator's control of the autonomous robot only extends to their general request for support, while the robot itself is responsible for conversational behaviors such as response generation and turn-taking. This makes it possible for the operator to control other robots within the conversation space while maintaining an acceptable cognitive load.

\section{Results}
We now describe the actual testing and implementation of the proposed system at the 2025 World Expo in Osaka, Japan. Note that due to privacy issues, we could not collect data at the expo itself, so no formal subjective evaluation was conducted. However we did conduct pilot testing of the speech recognition functionality in a controlled laboratory environment. 

\subsection{Pilot testing}
The speech recognition tests involved the same user speaking under several conditions. These conditions were:

\begin{itemize}
	\item The user holding a hand-held microphone (no robot)
	\item The user speaking at a stationary Teleco 
	\item The user speaking at a Teleco which was constantly rotating back and forth
	\item The user speaking at a Teleco with a large amount of simulated background noise coming from speakers
\end{itemize}

For conditions which involved a Teleco robot, the user spoke from a distance of approximately one meter, approximately the same that for casual conversation. We then measured the character error rate (CER) of the system under these conditions. Our findings are presented in Table \ref{results}.

\begin{table}[h]
\caption{CER of speech recognition under various conditions}
\label{results}
\begin{center}
\begin{tabular}{|c||c|}
\hline
Condition & CER (\%)\\
\hline
Hand-held microphone& 7.87\\
Stationary Teleco& 6.71\\
Moving Teleco& 6.71\\
Teleco in noisy environment & 7.18\\
\hline
\end{tabular}
\end{center}
\end{table}

We found that the CER of the speech recognition system is reasonably well-performing, and comparable to a hand-held microphone. Although we just used one user for these pilot tests, we also found similar performance for other users.

\subsection{Live expo demonstration}
The systems were run simultaneously in the same building (pavilion) for six days during the month of August. The Expo itself had many visitors and members of the public were invited to use the systems. The pavilion also contained multiple other robot demonstrations. Across six days, approximately 7300 people entered the pavilion. There were 513 interactions with ERICA (both pairs of people and individuals), however we did not log the number of interactions with the Teleco robots.

The pavilion environment contained a large amount of reverberation and high noise levels, with thousands of members of the public entering the pavilion on a daily basis. Background speech from other exhibits and occasionally heavy rainfall would create extra noise inside the pavilion. Our audio system was necessary to deal with this extreme environment. 

Our system was able to successfully recognize conversational speech in the majority of cases. Coherent and relevant responses from both ERICA and Teleco showed that conversation flow could be maintained in both scenarios. Monitoring of Teleco's audio system showed that selection of the appropriate audio channel based on the position and orientation of participants was successful. Several participants who took the role of the operator remarked that they could clearly hear the conversation despite the high level of noise. This shows that using enhanced audio channels to stream to operators rather than raw audio should be considered in similar public spaces.

\begin{figure}[t]
		\centering
		\includegraphics[width=\columnwidth]{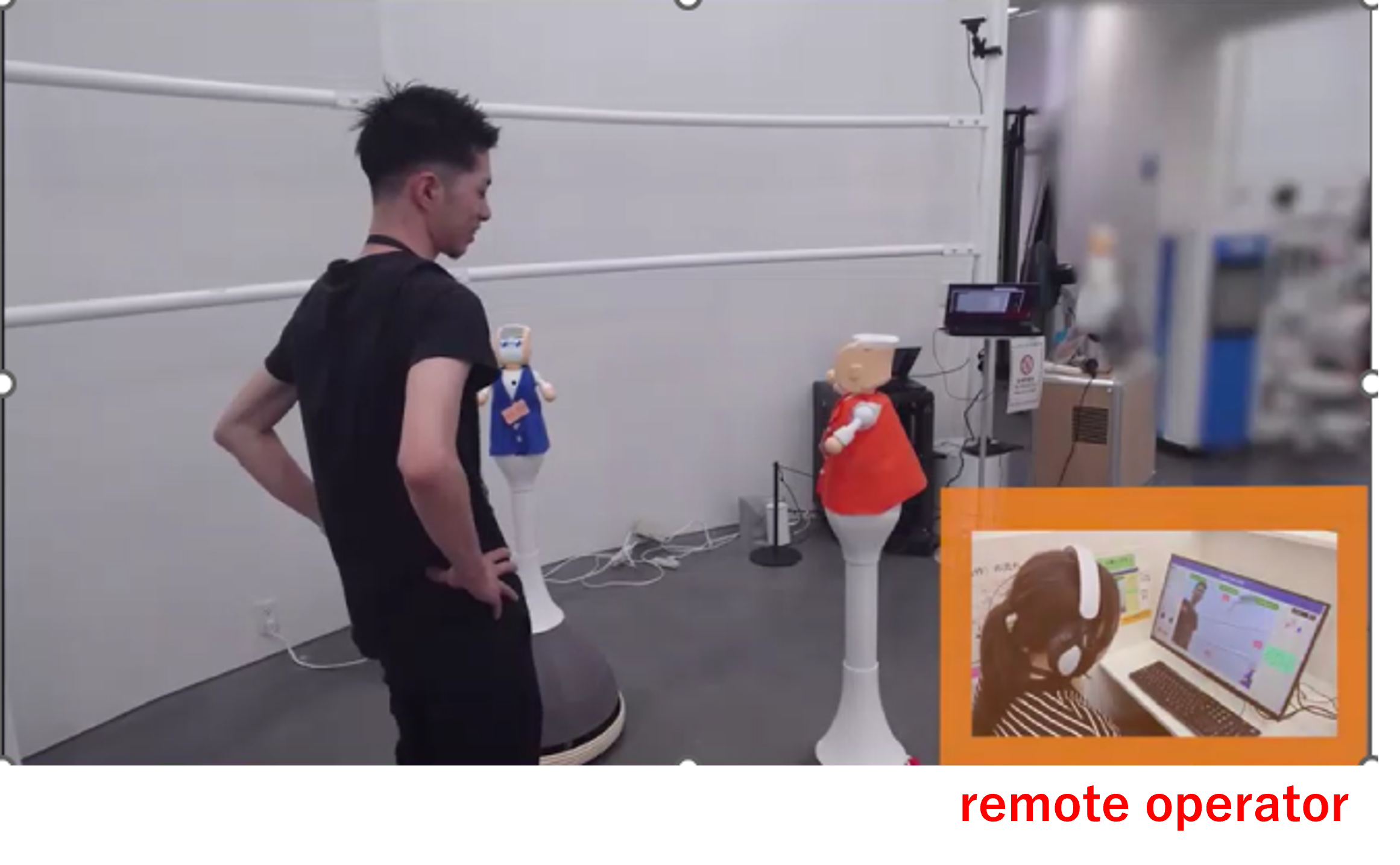}
		\caption{A participant engaging in a multi-party conversation with the two Teleco robots.} 
		\label{teleco_interaction}
 \end{figure}

\subsection{Limitations}
While the implementation was successful, there is room to improve the system. One major issue is that the dialogue system itself is not truly multi-party. In attentive listening, the participants interact with ERICA one at a time, while in the conversational support scenario the remote operator explicitly commands the robot to interject during the conversation. Multi-party conversation requires the autonomous robot to be able to engage in smooth turn-taking, in particular addressee detection. There have been attempts to use LLMs to solve this \cite{Hilgert2025,Addlesee2024b}, however this would probably require a multimodal approach using vision models.

Speech recognition was generally acceptable, although we observed situations, particularly in the conversational support scenario, when the user's speech was not picked up by the microphone array. We identified two main sources of this problem. The first is that the speaker's voice was too quiet to be recognized. The second issue was that some users would turn and talk to \textbf{T-AUTO}. This is a natural behavior, but it also meant their voice was not able to be processed correctly. Since \textbf{T-AUTO} would only produce dialogue when instructed to by the operator, it would remain silent and the conversation would stagnate.

\section{CONCLUSION}
This work described two multi-party conversational robot systems with different settings, but using audio processing based on the same microphone array system. The settings varied in terms of the distribution of humans and robots, the amount of mobility, and the location of the microphone. We showed that we could accommodate both these scenarios through speech enhancement and appropriate selection of microphone array audio channels. We demonstrated both systems as proofs of concept at the 2025 World Expo in Osaka, Japan and found that they were robust to noise and could facilitate multi-party conversation.

\section*{ACKNOWLEDGMENT}
This work was supported by JST Moonshot R\&D JPMJPS2011.

\bibliographystyle{IEEEtran}
\bibliography{ROMAN2026_bib}

\end{document}